\documentclass[journal]{IEEEtran}
\AtBeginDocument{%
  }
\usepackage{amsmath}

\usepackage{amsfonts,amssymb}
\usepackage[T1]{fontenc}% optional T1 font encoding
\usepackage{hyperref}
\hypersetup{
    colorlinks=true,
    linkcolor=blue,
    filecolor=blue,      
    urlcolor=blue,
    citecolor=cyan,
}
\usepackage{xcolor}
\usepackage{arydshln}

\ifCLASSINFOpdf
   \usepackage[pdftex]{graphicx}
\else
\fi
\usepackage{amsmath}

\usepackage[cmintegrals]{newtxmath}
\usepackage{bm}
\usepackage{algorithm}% http://ctan.org/pkg/algorithms
\usepackage{algorithmic}% http://ctan.org/pkg/algorithmicx
\usepackage{indentfirst}
\usepackage{amsmath,amssymb}
\usepackage{array}
\usepackage{mathrsfs}
\usepackage{cite}
\usepackage{tikz}
\usepackage{multirow} 
\usepackage{booktabs}
\usepackage{colortbl}
\usepackage{color}
\usepackage{comment}
\definecolor{mypurple}{RGB}{140,136,177}
\definecolor{myred}{RGB}{165,43,85}
\definecolor{mygreen}{RGB}{16,139,150}
\newcommand*{\circled}[1]{\lower.7ex\hbox{\tikz\draw (0pt, 0pt)%
    circle (.5em) node {\makebox[1em][c]{\small #1}};}}

\usepackage{soul}
\newcommand{\etal}{\textit{et al}.~}
\newcommand{\ie}{\textit{i}.\textit{e}.}
\newcommand{\eg}{\textit{e}.\textit{g}.}
\ifCLASSOPTIONcompsoc
 \usepackage[caption=false,font=normalsize,labelfont=sf,textfont=sf]{subfig}
\else
 \usepackage[caption=false,font=footnotesize]{subfig}
\fi
\usepackage{url}
\begin{document}

\title{Deep Shape-Texture Statistics for Completely Blind Image Quality Evaluation}

\author{Yixuan~Li,
        Peilin~Chen,
        Hanwei~Zhu,
        Keyan Ding, Leida~Li,~\IEEEmembership{Member,~IEEE},
   and~Shiqi~Wang,~\IEEEmembership{Senior Member,~IEEE}% <-this % stops a space
\thanks{Y. Li, P. Chen, H. Zhu, and S. Wang are with the Department of Computer and Science, City University of Hong Kong, Hong Kong (e-mail: yixuanli423@gmail.com; plchen3@cityu.edu.hk; hanwei.zhu@my.cityu.edu.hk; shiqwang@cityu.edu.hk). 

K. Ding is with the Hangzhou Global Scientific and Technological Innovation Center, Zhejiang University, Zhejiang, China (e-mail: dingkeyan@zju.edu.cn). 

L. Li is with the School of Artificial Intelligence, Xidian University, Xi'an, China (e-mail: reader1104@hotmail.com).

Corresponding author: Shiqi Wang.}% <-this % 
}

\markboth{Submitted to SUBMITTED TO xxxx}%
{Shell \MakeLowercase{\textit{et al.}}: Bare Demo of IEEEtran.cls for IEEE Communications Society Journals}
\maketitle
\begin{abstract}
Opinion-Unaware Blind Image Quality Assessment (OU-BIQA) models aim to predict image quality without training on reference images and subjective quality scores. Thereinto, image statistical comparison is a classic paradigm, while the performance is limited by the representation ability of visual descriptors. Deep features as visual descriptors have advanced IQA in recent research, but they are discovered to be highly texture-biased and lack of shape-bias. On this basis, we find out that image shape and texture cues respond differently towards distortions, and the absence of either one results in an incomplete image representation. Therefore, to formulate a well-round statistical description for images, we utilize the shape-biased and texture-biased deep features produced by Deep Neural Networks (DNNs) simultaneously. More specifically, we design a Shape-Texture Adaptive Fusion (STAF) module to merge shape and texture information, based on which we formulate quality-relevant image statistics. The perceptual quality is quantified by the variant Mahalanobis Distance between the inner and outer Shape-Texture Statistics (DSTS), wherein the inner and outer statistics  respectively describe the quality fingerprints of the distorted image and natural images. The proposed DSTS delicately utilizes shape-texture statistical relations between different data scales in the deep domain, and achieves state-of-the-art (SOTA) quality prediction performance on images with artificial and authentic distortions. 
\end{abstract}


% \received{xxxx April 2023}


\section{Introduction}
\IEEEPARstart{I}{n} the digital content boosting era, high-quality images are highly desired. However, images are inevitably corrupted by ubiquitous distortions during image formulation or after-processing. Quality perception is the natural instinct for the human visual system (HVS), but it is nontrivial to be modeled mathematically. Therefore, it is essential to design effective image quality measures for monitoring and evaluating the visual quality of digital images. 

Blind image quality assessment (BIQA) targets at mimicking human quality perception and automatically quantifies quality disturbance without information from corresponding pristine images. It is necessarily adaptable to wide application scenarios but is challenging due to the absence of reference information. Essentially, BIQA methods~\cite{BIQI,diivine,BLIINDS-II,brisque,lbiq,li2011blind,corina} follow the pipeline of ``\emph{Feature extraction-Supervised feature fusion-Quality prediction}'', where the mapping to the objective quality is learned under the supervision of mean opinion scores (MOSs), \eg, learning a regression model. In other words, these models are exposed to MOSs while training. However, MOSs are the subjective image quality labels that are singly annotated by crowd-sourcing, becoming costly and inefficient in today's data-boosting context. Such dependence on the supervision of MOSs may pose challenges in scalability and efficiency of quality prediction. Recognizing these challenges, the need for opinion-unaware (OU) BIQA methods is becoming increasingly pressing. Such OU-BIQA models offer a promising direction by eliminating the reliance on subjective opinions and reference images in model learning, paving the way for more robust and versatile image quality assessment in diverse applications.

\begin{figure}[!tbp]
\centering
\includegraphics[scale=0.88]{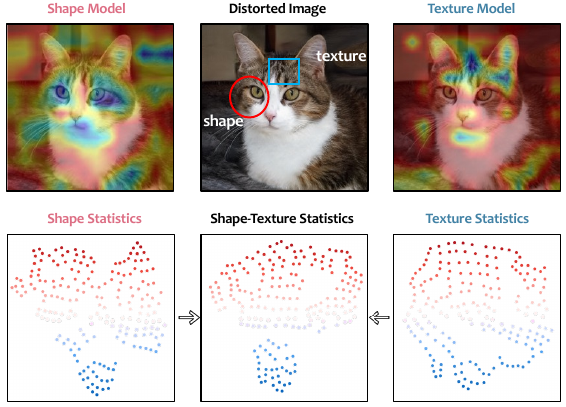}
\caption{Visualization of image shape-texture activation and statistical distribution. The top row contains a distorted image and its Class Activation Mapping (CAM) maps~\cite{selvaraju2017grad} of shape and texture models, which indicate the regions attended by models. Cooler color indicates more attentions paid by models. The second row contains the t-SNE~\cite{tsne} visualized image statistical distributions in terms of shape, texture, and shape-texture deep features.}
\label{probf}
\end{figure}
In contrast to opinion-aware (OA) BIQA methods, OU-BIQA models can assess the quality of an arbitrary image without training on MOSs, aiming at \emph{completely blind}. Current OU-BIQA methods can be classified into three types: natural scene statistics (NSS) based~\cite{mittal2011blind,niqe,ilniqe,snpniqe,babu2023no,shukla2024opinion}, pseudo-label assisted~\cite{wang2023toward,dipiq,bliss}, and neural network based~\cite{zhu2021recycling}. Thereinto, the NSS-based statistical comparison methods takes a preponderant role due to the sound NSS theories. NSS models demonstrate that natural images possess high statistical regularities~\cite{spatialnss,waveletnss,simoncelli2001natural,geusebroek2005six}, whose quality representation capability has been thoroughly investigated in different domains, such as spatial domain~\cite{niqe}, wavelet transform domain~\cite{diivine}, and discrete cosine transform domain~\cite{BLIINDS-II}. However, the performance of NSS-based methods still remains somewhat unsatisfactory in OU-BIQA, implying the projected coefficients may not sufficiently capture meaningful information denoting the intrinsic image quality. Meanwhile, IQA methods utilizing pretrained deep neural networks (DNNs) have gained prominence in recent years~\cite{amirshahi2016image,gao2017deepsim,berardino2017eigen,kim2017deep,lpips,liao2022deepwsd}, attributed to the powerful quality representation capability of deep learning features. Therefore, we tend to cultivate the statistical property of images within a more quality-relevant domain, \emph{i.e.}, the deep learning feature domain. However, the IQA-preferred off-the-shelf deep features produced by current pretrained DNNs are verified to be highly \emph{texture-biased}~\cite{geirhos2018imagenet,islam2021shape,mummadi2021does,naseer2021intriguing}, referring to the over-reliance on object texture compared with the shape. On the contrary, it is universally revealed that human cognition highly relies on shapes (e.g., silhouettes of cats) in images instead of textures (e.g., markings of fur)~\cite{landau1988importance,kucker2019reproducibility}. Accordingly, utilizing singly texture-biased deep features may not be able to represent image quality completely. The shape-biased and texture-biased features turn out to attend on complementary image areas (as shown in the first row of Fig.~\ref{probf}), where they accordingly exhibit different quality statistics that reflect different patterns of manifestation (as shown in the second row of Fig.~\ref{probf}), which can be employed to formulate a well-rounded shape-texture statistical description for image quality. Besides, the effectiveness of incorporating shape-relevant cues in IQA has been validated by previous works~\cite{zhang2017multiple,li2020quality,ding2021locally}. Therefore, in this paper, we incorporate \textit{shape-biased} with \textit{texture-biased} deep features which can compromise a more complete quality description of images. \\
\indent Herein, we make one of the first attempts to investigate the \emph{shape-texture} oriented image statistics, based on which we establish an NSS-based model for completely blind IQA. In specific, we utilize two DNN branches to respectively extract shape-biased and texture-biased deep features, and we design a Shape-Texture Adaptive Fusion (STAF) module to merge them together. Based on the shape-texture oriented deep representation, we formulate the inner statistics that denote the intrinsic quality patterns of each distorted image, and the outer statistics that represent the quality fingerprints of the natural image domain. The model predicts image quality by measuring the statistical Distance between the inner and outer Shape-Texture Statistics (\textbf{DSTS}). 
The proposed deep shape-texture statistical notion is quality-aware and free of task-specific training, enabling DSTS to be able to measure the perceptual quality effectively without reference images, MOSs, or any other additional information. We perform extensive experiments to demonstrate the superiority of the proposed DSTS. It turns out that DSTS has conspicuous performance in terms of quality prediction accuracy, generalization ability, and personalized quality assessment.

\section{Related Works}
\subsection{OU-BIQA Methods}
The pioneer work of OU-BIQA is proposed by Mittal \emph{et al.}~\cite{mittal2011blind}, where the quality-aware visual-word distributions are formulated to evaluate visual quality without subject opinions, but the distortion type as auxiliary information is still required. Later, in \cite{niqe}, a natural image quality evaluator (NIQE) is proposed based on the comparison of NSS features between a bunch of pristine images and the distorted image. Furthermore, Zhang \emph{et al.}~\cite{ilniqe} and Liu \emph{et al.}~\cite{snpniqe} improved the NIQE to a patch-level evaluation in terms of refined quality features. Xue \emph{et al.}~\cite{qac} addressed the opinion-unaware problem by learning quality-aware centroids as the codebook to predict image patch quality, while it can only deal with four distortion types. In \cite{pique}, local patch features are extracted and projected into quality scores according to unsupervised patch distortion classification. In \cite{bliss}, the authors proposed to train a model using pseudo opinion scores derived from full-reference IQA models, while in \cite{rankiqa} the pseudo quality rank is employed for supervision. Wu \emph{et al.}~\cite{lpsi} proposed an unsupervised effective local pattern statistics index (LPSI) by reforming local binary patterns into quality-aware features. Ma \emph{et al.}~\cite{dipiq} chose best-trusted IQA measures to form quality-discriminable image pairs and adopted the pair-wise learning-to-rank algorithm to learn the quality projection without subjective opinions. Zhu \emph{et al.}~\cite{zhu2021recycling} proposed to employ the discriminator of trained Wasserstein generative adversarial networks (WGAN) to measure the distance between pristine images and distorted image. Thereafter, Wang \etal~\cite{wang2023toward} established a new image dataset with pseudo quality labels, based on which they trained a DNN-based IQA model considering the domain shift caused by distortion and content between the distorted and natural images. Though numerous OU-BIQA models are proposed, the one that has better correlations to human visual quality is still in need.

\subsection{Shape and Texture Bias in CNNs}
ImageNet~\cite{deng2009imagenet}-pretrained CNNs exhibit excellent performance in many computer vision tasks~\cite{krizhevsky2012imagenet}. However, it has been found that they extensively rely on superficial textural cues rather than shapes, which runs counter to human vision~\cite{landau1988importance,kucker2019reproducibility}. Geirhos \emph{et al.}~\cite{geirhos2018imagenet} initially propounded CNNs' over-preference towards texture as \emph{texture-bias}. The authors consolidated the opposite bias between the recognition of human and CNNs using cue-conflict images (\textit{e.g.}, a zebra-textured cat). They proposed a CNN training strategy for better robustness by removing texture information in images with style transfer models.  Islam \emph{et al.}~\cite{islam2021shape} provided further evidence that texture bias appears not only in the outputs of CNNs, but also in the latent representations of intermediate layers. Thereafter, Mummadi \emph{et al.}~\cite{mummadi2021does} pointed out that solely increasing the shape-bias of CNNs can result in increased sensitivity towards distortions, which can conversely benefit the representation capability of image quality corruptions. Recently, Qiu \emph{et al.}~\cite{qiu2024shape} demonstrated that integrating shape and texture information yields superior out-of-distribution robustness compared to the adoption of either attribute in isolation. Therefore, incorporating shape and texture information simultaneously can formulate a unified shape-texture deep representation for images.

\subsection{Image Statistics}
Image statistics exhibit distinct properties and connotations across diverse scales of image space. According to data range, it can be classified into scene scale statistics, domain scale statistics, and single-image scale statistics.

\noindent\textbf{Scene Scale Statistics}\quad Image scene statistics refer to the general property possessed by mass natural images. For example, the NSS models reveal that the responses of natural images to band-pass filters obey the generalized Gaussian distribution (GGD) or Gaussian scale mixture model (GSM)\cite{simoncelli2001natural,spatialnss,waveletnss,geusebroek2005six}. 

\noindent\textbf{Domain Scale Statistics}\quad Domain-scale statistics reflect the shared characters in terms of content, attribute, or style of an image set, which are mostly gained by training DNNs on image databases~\cite{liu2015deep}. To this extent, the domain-scale image statistics derived from a set of homogeneous and representative images can be regarded as a sufficient estimation of scene-scale image statistics de facto. 

\noindent\textbf{Single Image Scale Statistics}\quad Single-image statistics represent the image internal statistical characteristics, which can be captured by analyzing the data distribution of its diverse local patches. In \cite{zhang2014image}, Zhang \emph{et al.} stated that the 3D transform coefficients of all image blocks within a single natural image together with those of their non-local similar blocks satisfy the GGD. In \cite{shaham2019singan} and \cite{shocher2019ingan}, GANs were adopted to learn the statistical patch distribution of a single image. 
\section{The proposed DSTS for OU-BIQA}
\subsection{Problem Formulation}
\label{problemformulation}
To predict the quality of distorted images, we first formulate image outer and inner statistical distributions and quantify visual quality with a quality-aware distance measure. In principle, the statistical distributions are based upon the shape-texture statistics. Herein, we denote the density $p(\mathbf{x})$ of the discrete image embedding $\varrho$ as the sum of Dirac delta distributions~\cite{heitz2021sliced}:
\begin{equation}
    p(\mathbf{x})=\frac{1}{N} \sum_{n=1}^{N} \delta_{ \varrho_{n}}(\mathbf{x}),
\label{eq3.1}
\end{equation}
where $x$ denotes the observation of the random variable, $N$ denotes the component number of $ \varrho$, and $\varrho_{n}$ is the $n\mbox{-}th$ component. 
We aim to define a quality-aware statistical distance measure between the inner and outer statistical distributions on the basis of the classic Mahalanobis Distance (MD)~\cite{bronstein2012taschenbuch}. As such, \ie, the larger the distance, the lower the visual quality. Suppose that the outer statistics obey the probability distribution with the mean value vector $\boldsymbol{\mu}_{G}$ and covariance matrix $\boldsymbol{\Sigma}_{G}$, meanwhile the inner statistics are with $\boldsymbol{\mu}_{M}$ and $\boldsymbol{\Sigma}_{M}$. Then their MD, denoted as $D_{m}(p_G (\mathbf{x}),p_M (\mathbf{x}))$, is given by,
\begin{equation}
\small
\begin{aligned}
    D_{m}(p_G (\mathbf{x}),p_M (&\mathbf{x})) = \\ 
    &\sqrt{(\boldsymbol{\mu}_{G}-\boldsymbol{\mu}_{M})^\intercal (\frac{\boldsymbol{\Sigma}_{G}+\boldsymbol{\Sigma}_{M}}{2})^{-1}(\boldsymbol{\mu}_{G}-\boldsymbol{\mu}_{M})}\ ,
\label{mdistance1}
\end{aligned}
\end{equation}
However, MD is an effective statistical distance metric mathematically yet remains limited in quantifying visual quality, because MD is position-agnostic in terms of the location relations between image patches~\cite{heitz2021sliced}, which are crucial for visual perception and content comprehension. 

\begin{figure*}[!tbp]
\centering
\includegraphics[scale=1.45]{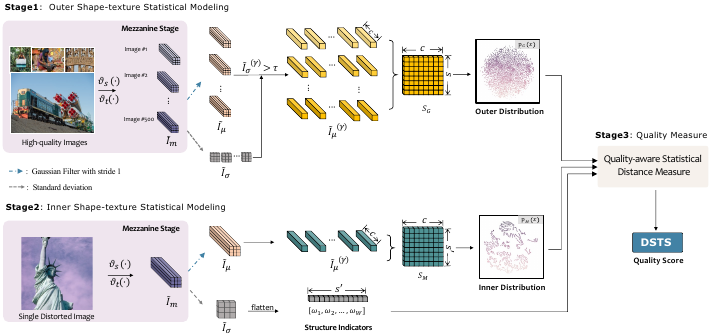}
\caption{Overview of the proposed DSTS framework. More specifically, DSTS contains three stages. In stage 1, we first formulate the image outer statistics based on a set of ideally pristine images with shape-texture statistics in deep domain. In stage 2, the inner statistics are extracted from each distorted image with shape-texture statistics in deep domain. In particular, in stages 1\&2, the mezzanine stage transforms the images into the perceptual space, which is detailed in Fig.~\ref{framework2}. In stage 3, the quality-aware statistical distance measure between the inner and outer distributions quantifies the perceptual quality.}
\label{framework1}
\end{figure*}

Therefore, we approximate the inner distribution with the patch samples, and the MDs between inner samples and the outer distribution are finally merged, leading to the final quality-aware distance, 
\begin{equation}
\small
\label{distance-m1}
    D_{q}(p_G (\mathbf{x}),p_M (\mathbf{x})) = \mathcal{F}(\{D_{m}(p_G (\mathbf{x}),\boldsymbol{M}_i)\}),\ i\in[1,W].
\end{equation}
Specifically, for the sample space $\boldsymbol{S}_M = \{\boldsymbol{M}_1, \boldsymbol{M}_2, \ldots, \boldsymbol{M}_W\}
$ of $p_M (\mathbf{x})$, the MD between each sample $\boldsymbol{M}_i$ and $p_G (\mathbf{x})$ is
\begin{equation}
\small
\begin{aligned}
D_{m}(p_G (\mathbf{x}),\boldsymbol{M}_i) = 
\sqrt{(\boldsymbol{\mu}_{G}-\boldsymbol{M}_{i})^\intercal (\frac{\boldsymbol{\Sigma}_{G}+\boldsymbol{\Sigma}_{M}}{2})^{-1}(\boldsymbol{\mu}_{G}-\boldsymbol{M}_{i})}. 
\end{aligned}
\label{distance-m}
\end{equation}
It is trivial to see that $D_{m}(p_G (\mathbf{x}),p_M (\mathbf{x}))$ can be represented as the linear combination of $\{D_{m}(p_G (\mathbf{x}),M_{i})\}$, $i\in[1, W]$. As such, by adjusting the weights according to the perceptual importance of each sample~\cite{ilniqe}, we can formulate a new quality-aware statistical distance measure:
\begin{align}
    D_{q}(p_G (\mathbf{x}),p_M (\mathbf{x})) 
    =\sum_{i=1}^W \omega_i \cdot D_{m}(p_G (\mathbf{x}),M_{i}),
\label{distance-q}
\end{align}
where $\omega_i$ is the carefully designed content weighting with the constraint of $\sum_{i=1}^W \omega_i=1$, and $W$ is the number of sample partitions. In this case, the $\mathcal{F}$ in Eqn.~(\ref{distance-m1}) is a linear combination. 
According to the problem formulations above, the proposed framework is crystallized. 
Overall, the pipeline is illustrated in Fig.~\ref{framework1}. 

\subsection{Shape-Texture Statistics Modeling}
In this subsection, we mathematically extract the inner and outer shape-texture statistics. We first present how the image is transformed into the shape-texture oriented perceptual domain represented with deep features, and then we establish the statistical modeling of both inner and outer statistics. 
\begin{figure}[!tbp]
\centering
\includegraphics[scale=0.95]{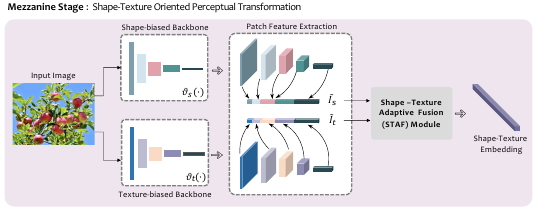}
\caption{Illustration of the mezzanine stage for shape-texture oriented perceptual transformation based on the shape and texture CNN branches. The final shape-texture embedding are composed of five stages of convolutional outputs fused by the proposed STAF module. }
\label{framework2}
\end{figure}

\noindent\textbf{Shape-Texture Oriented Perceptual Transformation} 
Deep features of pretrained classification networks have been repeatedly shown to efficiently represent the rich visual information from the raw pixel domain, which is crucial for characterizing perceptual distortions~\cite{lpips}. Along this vein, we adopt the shape-biased and texture-biased EfficientNet-b7~\cite{tan2019efficientnet} simultaneously as transformation backbones, denoted as $\vartheta_{s}(\cdot)$ and $\vartheta_{t}(\cdot)$ respectively. The overall perceptual transformation process is illustrated in Fig.~\ref{framework2}. Given an arbitrary image $\boldsymbol{I}$, the transformation can be obtained by $\tilde{\boldsymbol{I}}_{s} = \vartheta_{s}(\boldsymbol{I})$ and $\tilde{\boldsymbol{I}}_{t} = \vartheta_{t}(\boldsymbol{I})$. We employ multiple-layer latent features simultaneously to capture size-dependent visual patterns contained in image patches~\cite{kligvasser2021deep}. In total, five layers of outputs are included respectively for the two substrates, which correspond to the $2^{nd}$ to $6^{th}$ convolutional stages. To fuse them, we iteratively conduct spatial down-sampling on the first to forth layers of outputs for four to one times respectively, using the 2-D low-pass filter $\boldsymbol{\psi}(\cdot)$ with the stride of 2, where the $\boldsymbol{\psi}(\cdot)$ specified by $\{r_{k,l}|k=-K,...,K;l=-L,...,L\}$ is a unit-volume circularly symmetric Gaussian window. Finally, the five components are sequentially concatenated along channels. As such, we can obtain the shape-biased and texture-biased deep features $\boldsymbol{\tilde{I}}_s$ and $\boldsymbol{\tilde{I}}_t$ respectively. 

To merge the shape and texture biased features, we propose a Shape-Texture Adaptive Fusion (STAF) module, to obtain the shape-texture deep embedding, denoted as  $\boldsymbol{\tilde{I}}_m = STAF(\boldsymbol{\tilde{I}}_s,\boldsymbol{\tilde{I}}_t)$.
In specific, we first compute the variance along the channel dimension at each spatial locations,
\begin{equation}
    \boldsymbol{V}_{s}(p,q) = var(\boldsymbol{\tilde{I}}_s(p,q)),\
    \boldsymbol{V}_{t}(p,q) = var(\boldsymbol{\tilde{I}}_t(p,q)),
\end{equation}
where $(p,q)$ denotes the spatial indices, $p \in \{1,2,...,P\}$, $q \in \{1,2,...,Q\}$, and $P$ and $Q$ denote feature's spatial dimensions. The $var(\cdot)$ denotes the computation of variance. We then calculate the shape-biased and texture-biased attentions respectively:
\begin{equation}
\begin{aligned}
    &\boldsymbol{A}_{s}(p,q) = \frac{\boldsymbol{V}_{s}(p,q)}{\boldsymbol{V}_{s}(p,q)+\boldsymbol{V}_{t}(p,q)},\\
    &\boldsymbol{A}_{t}(p,q) = \frac{\boldsymbol{V}_{t}(p,q)}{\boldsymbol{V}_{s}(p,q)+\boldsymbol{V}_{t}(p,q)}.
\end{aligned}
\end{equation}
The final shape-texture embedding is obtained by adaptively aggregating $\boldsymbol{\tilde{I}}_s$ and $\boldsymbol{\tilde{I}}_t$ together:
\begin{equation}
    \boldsymbol{\tilde{I}}_m = \boldsymbol{\tilde{I}}_s \odot \boldsymbol{A}_{s} + \boldsymbol{\tilde{I}}_t \odot \boldsymbol{A}_{t},
\end{equation}
where $\odot$ denotes the Hadamard product.
In this case, an image is transformed into the shape-texture oriented perceptual domain.

\noindent\textbf{Statistical Modeling} 
Generally speaking, image local statistical descriptions could work as efficient visual pattern descriptors~\cite{julesz1962visual}. Therefore, we first compute the intra-channel local average maps $\boldsymbol{I}_{\mu}$ and standard deviation maps $\boldsymbol{I}_{\sigma}$ of $\boldsymbol{\tilde{I}}_m$ respectively:
\begin{equation}
    \boldsymbol{I}_{\mu}(p,q) = \sum^{K}_{k=-K}\sum^{L}_{l=-L} r_{k,l}\cdot \boldsymbol{\tilde{I}}_{m}(p+k,q+l),
\end{equation}
\begin{equation}
    \boldsymbol{I}_{\sigma}(p,q) = \sqrt{\sum^{K}_{k=-K}\sum^{L}_{l=-L}  r_{k,l}\cdot (\tilde{\boldsymbol{I}}_{m}(p+k,q+l)-\boldsymbol{I}_{\mu}(p,q))^2}.
\end{equation}
To make the statistical analysis across different convolutional layers with uncertain magnitude ranges consistent, we conduct the layer-wise $\ell_2$-normalization to project $\boldsymbol{I}_{\mu}$ of convolutional layer $i$ onto the unit hypersphere using
\begin{equation}
    \boldsymbol{\tilde{I}}_{\mu i} = \frac{\boldsymbol{I}_{\mu i}}{{||\boldsymbol{I}_{\mu i}||}_{2}},
\end{equation}
and then all the $\tilde{\boldsymbol{I}}_{\mu i}$ components are concatenated into $\boldsymbol{\tilde{I}}_{\mu}$ sequentially. We utilize $\gamma$ to denote $(p,q)$ for simplification, where $\gamma \in \{1,W\}$ and $W$ is the number of spatial patches.

In DSTS, the component $\boldsymbol{I}_{\mu}^{(\gamma)}$ at each spatial location $\gamma$ is treated as a random observation sampled from the inner and outer statistical distributions. These distributions are multivariate with respect to the dimension of feature channel number $c$. For the image outer statistics, it requires an adequate amount of candidate images as the knowledge source to describe the statistical characteristics of  images. However, not all observations are entirely equal in representing the quality of the natural image domain. We observe that heterogeneous image areas react more sensibly to distortion disturbance compared with homogeneous areas, implying that they can reflect more quality-relevant cues. Therefore, we only consider the samples containing vital image structures (\eg, edges, corners). To distinguish these patches apart, we compute the cross-channel average of the deviation $\boldsymbol{I}_{\sigma}$ to form a structure indicator:
\begin{equation}
\boldsymbol{\tilde{I}}_{\sigma}^{(\gamma)} = \frac{1}{c}\sum_{o=1}^{c}\boldsymbol{I}_{\sigma o}^{(\gamma)},
\end{equation}
where $o$ is the channel index. We only consider the patches with $\boldsymbol{\tilde{I}}_{\sigma}^{(\gamma)} \geq \tau$, where $\tau$ is the filtering threshold. Then we stack all qualified observations derived from all the \emph{good-quality} images into the sample set $\boldsymbol{S}_G \in \mathbb{R}^{s\times c}$, where $s$ denotes the number of qualified observations from all pristine images.
The outer statistics are characterized by the mean and covariance of $\boldsymbol{S}_G$, denoted as $\boldsymbol{\mu}_G$ and $\boldsymbol{\Sigma}_G$:
\begin{equation}
    \boldsymbol{\mu}_{G}(1,j)=\frac{1}{s}\sum^{s}_{i=1}\boldsymbol{S}_{G}(i,j) ,
\end{equation}
\begin{equation}
    {\boldsymbol{\Sigma}_G}(i,j) =\frac{1}{s-1}(\boldsymbol{S}_G -{\boldsymbol{\mu}_{G}})^\intercal(\boldsymbol{S}_G -{\boldsymbol{\mu}_{G}}),
\end{equation}
where $i,\ j$ denote element indice in $\boldsymbol{S}_G$, $\boldsymbol{\mu}_{G}\in \mathbb{R}^{1\times c}$, and ${\boldsymbol{\Sigma}_G}\in \mathbb{R}^{c \times c}$.

Meanwhile, the image inner statistics aim at capturing the intrinsic quality pattern contained in a single image. Analogously, we count $\boldsymbol{\tilde{I}}_{\mu}^{(\gamma)}$ of the distorted image as a random observation of the inner distribution, and stack into $\boldsymbol{S}_M \in \mathbb{R}^{{s}' \times c}$, where ${s}'$ denotes the number of observations in the distorted image. In this case, all the samples are from the same distorted image without filtering. Similarly, the inner statistics are specified with the mean and covariance of $\boldsymbol{S}_M$, represented by $\boldsymbol{\mu}_M$ and $\boldsymbol{\Sigma}_M$.

\subsection{DSTS: Distance Based on Shape-Texture Statistics}
In Section~\ref{problemformulation}, we have specified the quality-aware distance measure between the inner and outer statistical distributions $p_M (\mathbf{x})$ and $p_G (\mathbf{x})$ on the basis of Mahalanobis distance (as shown in Eqn.~(\ref{distance-m}) and Eqn.~(\ref{distance-q})). 
It is worth mentioning that the outer statistics are obtained before the quality prediction stage.
As such, DSTS is a completely blind image quality indicator de facto, where each distorted image is insulated from its pristine counterpart and quality label. 
To ensure the covariance matrices $\boldsymbol{\Sigma}_G$ and $\boldsymbol{\Sigma}_M$ positive definite, a regularization term $\lambda$ is added to the diagonal elements:
\begin{equation}
    \boldsymbol{\Sigma}_G = \boldsymbol{\Sigma}_G + \lambda \cdot I_0, \ 
    \boldsymbol{\Sigma}_M = \boldsymbol{\Sigma}_M + \lambda \cdot I_0,
\end{equation}
where $I_0$ is the identity matrix, and $\lambda = 1\times e^{-6}$. The DSTS index of each distorted image is computed with
\begin{equation}
\small
    \begin{aligned}
        DS&TS = D_q(p_G(\mathbf{x}),p_M(\mathbf{x}))=\\
        &\sum ^{W}_{\gamma=1}\boldsymbol{\tilde{I}}_{\sigma}^{(\gamma)}\cdot \sqrt{(\boldsymbol{\mu}_G-\boldsymbol{\tilde{I}}_{\mu}^{(\gamma)})^{\intercal}(\frac{\boldsymbol{\Sigma}_G + \boldsymbol{\Sigma}_M}{2})^{-1}(\boldsymbol{\mu}_G-\boldsymbol{\tilde{I}}_{\mu}^{(\gamma)})},
    \end{aligned}
\end{equation}
where $\boldsymbol{\tilde{I}}_{\sigma}^{(\gamma)}$ serves as the content weighting parameter in Eqn.~(\ref{distance-q}). 
Therefore, a larger DSTS value indicates a lower quality level. 

\subsection{Connections to Relevant IQA Methods}
The proposed DSTS scheme is closely related to several IQA models. Herein, we elaborate on the connections and differences.

\noindent\textbf{NIQE and its variants}: The NIQE~\cite{niqe} and its variants~\cite{ilniqe,snpniqe,npqi} have been proposed to build NSS models based on low-level visual descriptors of images, where the NSS fitting parameters are extracted and utilized as the to-be-compared quality features. Our model partially inherits this paradigm. Specifically, instead of extracting handcrafted NSS features from the  low-level visual descriptors, we directly employ the shape-texture statistics based on deep features to achieve statistical comparisons, which exploits more visually relevant quality cues.

\noindent\textbf{Deep Statistical Comparisons based FR-IQA}: 
The statistical comparison FR-IQA models aim to model the HVS by comparing feature statistics~\cite{duanmu2021quantifying}. In~\cite{gatys2015texture,rabin2012wasserstein,kolkin2019style}, feature distribution comparisons have been verified to be effective in image quality optimization. Deep Wasserstein distance~\cite{liao2022deepwsd} calculates the visual quality based on the 1-D Wasserstein distance between the statistical distributions of deep features derived from the distorted and reference image patches. Deep distance correlation~\cite{zhu2022distance} measures both linear and non-linear correlations in the deep feature domain. Although they both make comparisons between the intermediate responses of DNN features, the effectiveness of such statistical comparisons highly relies on the pristine reference images for FR-IQA. In contrast, DSTS extracts and compares quality-relevant statistics in respect of shape and texture without the necessity to align by semantic cues or contents. 

\section{Experiments}
In this section, we first provide the experimental setups, and then extensively verify the effectiveness of the proposed DSTS regarding image quality prediction accuracy on 13 IQA databases with a broad spectrum of distortions. Moreover, we demonstrate DSTS's superiority on generalization ability and personalized quality assessment even compared with OA-BIQA models. 
\subsection{Experimental Settings} 
\subsubsection{IQA Databases}
To validate the performance of the proposed DSTS, we evaluate it on (1) five synthetic distortion IQA databases: LIVE~\cite{live}, CSIQ~\cite{csiq}, TID2013~\cite{tid2013}, KADID-10k~\cite{kadid}, MDIVL~\cite{mdivl}, and MDID~\cite{mdid}; (2) five authentic distortion IQA databases: KonIQ-10k~\cite{koniq}, CLIVE~\cite{clive}, BID~\cite{bid}, CID2013~\cite{cid2013}, and SPAQ~\cite{spaq}; and (3) three databases containing generative distortions: AGIQA-3k~\cite{li2023agiqa}, GFIQA-20k~\cite{su2023going}, and LGIQA~\cite{gu2020giqa} (containing cityscapes, cat and ffhq subsets). The database details are listed in the supplementary materials (SM).
\subsubsection{Evaluation Criteria}
Three criteria are employed to evaluate performance, including Pearson Linear Correlation Coefficient (PLCC), Root Mean Square Error (RMSE), and Spearman Rank order Correlation Coefficient (SRCC). In particular, the PLCC and RMSE are responsible for prediction accuracy, while SRCC can reflect the monotonicity of prediction. The IQA method contains superiority when it has higher PLCC, SRCC, and lower RMSE. The PLCC and RMSE are computed after a five-parameter non-linear mapping between subjective and objective scores using the function as follows:
\begin{equation}
    f(x)=a_{1}\left\{\frac{1}{2}-\frac{1}{1+exp[{a_{2}(x-a_{3})]}}\right\}+a_{4}x +a_{5},
    \label{5fitting}
\end{equation}
where $x$ and $f(x)$ respectively denotes the predicted and mapped score, and $a_1$ to $a_5$ are fitting parameters for logistic regression.
\begin{table*}[htbp]
  \fontsize{8}{9}\selectfont
  \centering
  \caption{Quality Prediction Performance Comparisons on Individual Database Containing Artificial Distortions. The First and Second Best Performance are Highlighted in \textbf{Bold} and \underline{Underline} Respectively. Cases Where Shape Outperforms Texture are Marked in {\color{myred}Red} and Cases Where Texture Outperforms Shape are Marked in {\color{mygreen}Green}. }
  \renewcommand{\arraystretch}{1.2}
    \setlength{\tabcolsep}{0.5mm}{\begin{tabular}{ccccccccccccccccccc}
\toprule   
\multirow{2}{*}{OU-IQA Methods} & \multicolumn{3}{c}{LIVE} & \multicolumn{3}{c}{CSIQ} & \multicolumn{3}{c}{TID2013} & \multicolumn{3}{c}{KADID} &
\multicolumn{3}{c}{MDID} &
\multicolumn{3}{c}{MDIVL} \\
\cmidrule(r){2-4}  \cmidrule(r){5-7} \cmidrule(r){8-10} \cmidrule(r){11-13}  \cmidrule(r){14-16}  \cmidrule(r){17-19}
 \multicolumn{1}{c}{}  &PLCC &SRCC  &RMSE &PLCC &SRCC  &RMSE &PLCC &SRCC  &RMSE &PLCC &SRCC  &RMSE &PLCC &SRCC  &RMSE&PLCC &SRCC  &RMSE\\
\midrule   
    NIQE  &0.902 & 0.906 &11.779 &0.693& 0.619 &0.189 &0.398 & 0.311 &1.137 &0.438& 0.378 &0.974&0.670 & 0.649 &1.636 &0.557 & 0.566 &19.829\\
    QAC   &0.807 & 0.868 &27.322 &0.593 & 0.480  &0.263 &0.437 & 0.372 &1.115 &0.390 & 0.239 &0.997 &0.489 & 0.324 &1.922 &0.571 & 0.552 &19.604\\
    PIQUE  &0.836 &0.840  &14.994 &0.675& 0.512 &0.194 &0.525 & 0.364 &1.055 &0.373 & 0.237 &1.005 &0.333 & 0.253 &2.078 &0.504 & 0.492 &20.625\\
    LPSI  &0.786 &0.818 &27.322 &0.695 & 0.522 &0.263 &0.489 & 0.395 &1.081 &0.367 & 0.148 &1.007 &0.354 & 0.031 &2.061 &0.559 & 0.574 &19.808\\
    ILNIQE &0.897 &0.898 &12.072 &\textbf{0.837} & \textbf{0.805} &\textbf{0.144} &0.588 & 0.494 &1.003 &0.575 & 0.541 &0.886 &0.724 & 0.690 &1.519 & 0.614 & 0.624 &18.845\\
    dipIQ &\textbf{0.933} &\textbf{0.938} &\textbf{9.823} &0.742 & 0.519 &0.176 &0.477 & 0.438 &1.090 &0.399 & 0.298 &0.993 &0.674 & 0.661 &1.628 &0.761 & 0.713 &15.498\\
    SNP-NIQE &0.899 & 0.907 &11.962 &0.702 & 0.609 &0.187 &0.432 & 0.333 &1.118 &0.442 &0.372 &0.971 &0.750 & 0.726&1.456 &0.626 & 0.625 &18.620\\
    NPQI  &0.915 & 0.911 &11.018 &0.725 & 0.634 &0.181 &0.454& 0.281  &1.104 &0.450 & 0.391 &0.967 &0.731 & 0.698 &1.503 &0.594 & 0.614 &19.214\\
    ContentSep &0.741 &0.748 &18.362 &0.363 &0.587 &0.245&0.221&0.253&1.209&0.464&0.506&0.960&0.329&0.403&1.987&0.239&0.164&23.188 \\
    \midrule
    DSS-VGG16 (Ours)  &0.828&0.840&15.325&0.747&\color{myred}0.717&0.175&\color{myred}0.574&\color{myred}0.443&\color{myred}1.016&\color{myred}0.576&\color{myred}0.561&\color{myred}0.885 &0.742&0.712&1.476&0.577&0.567&19.567\\
    DTS-VGG16 (Ours)  &\color{mygreen}0.893&\color{mygreen}0.896&\color{mygreen}12.314&\color{mygreen}0.762&0.711&\color{mygreen}0.170&0.563&0.431&1.024&0.546&0.527&0.907&0.765&0.742&1.419&\color{mygreen}0.729&\color{mygreen}0.722&\color{mygreen}16.478\\
    DSTS-VGG16 (Ours) &0.838&0.881&13.303&0.765&0.720&0.169&0.588&0.445&1.003&0.573&0.552&0.888 &0.768&0.751&1.412&0.667&0.664&17.803\\
    \cdashline{1-19}
    DSS-ResNet50 (Ours)  &0.859&0.869&14.006&\color{myred}0.774&\color{myred}0.743&\color{myred}0.169&\color{myred}0.569&\color{myred}0.473&\color{myred}0.993&\color{myred}0.588&\color{myred}0.567&\color{myred}0.875&0.716&0.697&1.534&0.632&0.628&18.504\\
    DTS-ResNet50 (Ours)  &\color{mygreen}0.872&\color{mygreen}0.879&\color{mygreen}13.358&0.770&0.732&0.167&0.542&0.456&1.042&0.559&0.542&0.878&\color{mygreen}0.744&\color{mygreen}0.728&\color{mygreen}1.472&\color{mygreen}0.783&\color{mygreen}0.781&\color{mygreen}14.849\\
    DSTS-ResNet50 (Ours)  &0.868&0.878&13.586&0.775&0.742&0.166&0.593&0.476&0.999&\underline{0.596}&0.574&0.869&0.753&0.739&1.449&0.694&0.693&17.189\\
    \cdashline{1-19}
    DSS (Ours) &0.868&0.880&13.560&0.787&0.770&0.162&\underline{0.622}&0.511&0.971&\color{myred}0.594&0.567&\color{myred}\underline{0.871}&0.773&0.747&1.398&0.641&0.643&18.332\\
    DTS (Ours) &\color{mygreen}\underline{0.932} &\color{mygreen}\underline{0.936} &\color{mygreen}\underline{9.890} &\color{mygreen}0.791 & \color{mygreen}0.777 &\color{mygreen}0.161 &\color{mygreen}\textbf{0.624} &\color{mygreen}\underline{0.536} &\color{mygreen}\underline{0.969} &0.594 & \color{mygreen}\underline{0.598} &0.871 &\color{mygreen}\textbf{0.807} &\color{mygreen}\underline{0.758} &\color{mygreen}\textbf{1.300} &\color{mygreen}\underline{0.792} & \color{mygreen}\underline{0.787}&\color{mygreen}\underline{14.578}\\
    \rowcolor{mypurple!20}\textbf{DSTS (Ours)} &0.931&0.935&9.954&\underline{0.823}&\underline{0.798}&\underline{0.155}&0.610&\textbf{0.538}&\textbf{0.936}&\textbf{0.637}&\textbf{0.622}&\textbf{0.820}&\underline{0.806}&\textbf{0.795}&\underline{1.304}&\textbf{0.799}&\textbf{0.803}&\textbf{14.137} \\
    \bottomrule
    \end{tabular}%
    }
  \label{result1}%
\end{table*}%
\begin{table*}[htbp]
  \fontsize{8}{9}\selectfont
  \centering
  \caption{Quality Prediction Performance Comparisons on Individual Database Containing Authentic Distortions. The First and Second Best Performance are Highlighted in \textbf{Bold} and \underline{Underline} Respectively. Cases Where Shape Outperforms Texture are Marked in {\color{myred}Red} and Cases Where Texture Outperforms Shape are Marked in {\color{mygreen}Green}.}
    \setlength{\tabcolsep}{1.2mm}{\begin{tabular}{cccccccccccccccc}
\toprule   
\multirow{2}{*}{OU-IQA Methods} & \multicolumn{3}{c}{KonIQ} & \multicolumn{3}{c}{CLIVE} & \multicolumn{3}{c}{BID} & \multicolumn{3}{c}{CID2013} &\multicolumn{3}{c}{SPAQ}\\
\cmidrule(r){2-4}  \cmidrule(r){5-7} \cmidrule(r){8-10} \cmidrule(r){11-13} \cmidrule(r){14-16}   
 \multicolumn{1}{c}{}  &PLCC &SRCC  &RMSE &PLCC &SRCC  &RMSE &PLCC &SRCC  &RMSE &PLCC &SRCC  &RMSE &PLCC &SRCC  &RMSE\\
\midrule   
    NIQE   &0.538& 0.530  &0.466 &0.494 & 0.450  &17.648 &0.461 & 0.458 &1.111 &0.670 & 0.659 &16.798 &0.712 & 0.703 &14.675 \\
    QAC    &0.291 & 0.340  &0.552 &0.211 & 0.046 &19.839 &0.290 & 0.300   &1.252 &0.138 & 0.030  &22.424 &0.023 & 0.092 &20.901 \\
    PIQUE   &0.296& 0.245 &0.528 &0.279 & 0.108 &19.490 &0.112 & 0.042 &1.244 &0.092 & 0.045 &22.547 &0.263 & 0.097 &20.165 \\
    LPSI   &0.106 & 0.224 &0.552 &0.299 & 0.083 &19.368 &0.099 & 0.043 &1.246 &0.436 & 0.323 &20.376 &0.274 & 0.068 &20.103 \\
    ILNIQE  &0.531 & 0.506 &0.468 &0.503 & 0.439 &17.538 &0.507 & 0.494 &1.079 &0.427 & 0.306 &20.476 &0.721 & 0.714 &14.491 \\
    dipIQ  &0.435 & 0.238 &0.497 &0.304 & 0.177 &19.338 &0.218 & 0.019 &1.222 &0.355 & 0.103 &21.170 &0.497 & 0.388  &18.134\\
    SNP-NIQE  &0.640 & 0.628 &0.424 &0.520 & 0.465 &17.343 &0.437 & 0.425 &1.126 &0.726 & 0.716 &15.570 &0.746 & 0.739 &13.912 \\
    NPQI   &0.615& 0.612 &0.436 &0.492 & 0.475 &17.673 &0.460 & 0.468 &1.112 &0.777 &0.770  &14.252 &0.675 & 0.674 &15.419  \\
    ContentSep &0.628&0.388&0.640&0.523&0.506&17.131&0.426&0.411&1.165&0.632&0.611&17.567&0.711&0.708&14.699 \\
    \midrule
    DSS-VGG16 (Ours)  &0.720&0.712&0.384&\color{myred}0.519&\color{myred}0.465&\color{myred}17.351&0.457&0.429&1.114&0.808&0.799&13.339&\color{myred}0.787&\color{myred}0.784&\color{myred}12.902\\
    DTS-VGG16 (Ours)  &\color{mygreen}0.739&\color{mygreen}0.729&\color{mygreen}0.372&0.473&0.413&17.888&\color{mygreen}0.518&\color{mygreen}0.486&\color{mygreen}1.071&\color{mygreen}0.836&\color{mygreen}0.827&\color{mygreen}12.439&0.772&0.767&13.086\\
    DSTS-VGG16 (Ours) &\underline{0.743}&0.730&\underline{0.370}&0.490&0.433&17.693&0.484&0.455&1.095&0.832&0.823&12.572&\underline{0.792}&\underline{0.789}&\underline{12.757}\\
    \cdashline{1-16}
    DSS-ResNet50 (Ours)  &\color{myred}0.666&\color{myred}0.669&\color{myred}0.412&\color{myred}0.459&\color{myred}0.409&\color{myred}18.029&0.396&0.354&1.149&\color{myred}0.856&\color{myred}0.846&\color{myred}\textbf{11.080}&\color{myred}0.760&\color{myred}0.758&\color{myred}13.577\\
    DTS-ResNet50 (Ours)  &0.641&0.636&0.424&0.392&0.339&18.676&\color{mygreen}0.467&\color{mygreen}0.430&\color{mygreen}1.107&0.852&0.846&11.872&0.735&0.732&14.176\\
    DSTS-ResNet (Ours) &0.670&0.672&0.410&0.446&0.392&18.168&0.429&0.389&1.131&0.852&0.847&11.865&0.764&0.761&13.496\\
    \cdashline{1-16}
    DSS (Ours) &0.673&0.676&0.408&0.506&0.457&17.509&0.479&0.485&1.099&0.793&0.779&13.792&\color{myred}0.789&\color{myred}0.786&\color{myred}12.839\\
    DTS (Ours) &\color{mygreen}0.712 & \color{mygreen}\underline{0.733} &\color{mygreen}0.388 &\color{mygreen}\underline{0.536}&\color{mygreen}\underline{0.482} &\color{mygreen}\underline{17.131} &\color{mygreen}\textbf{0.579}&\color{mygreen}\textbf{0.565} &\color{mygreen}\textbf{1.022} &\color{mygreen}\underline{0.869} & \color{mygreen}\underline{0.853} &\color{mygreen}11.218 &0.747 &0.741 &13.889\\
    \rowcolor{mypurple!20}\textbf{DSTS (Ours)} &\textbf{0.752}&\textbf{0.753}&\textbf{0.370}&\textbf{0.539}&\textbf{0.483}&\textbf{17.101}&\underline{0.577}&\underline{0.565}&\underline{1.023}&\textbf{0.875}&\textbf{0.860}&\underline{11.108}&\textbf{0.806}&\textbf{0.801}&\textbf{12.383} \\
    \bottomrule
    \end{tabular}%
    }
  \label{result2}%
\end{table*}%
\begin{figure*}[!htbp]
\centering
\includegraphics[scale=0.7]{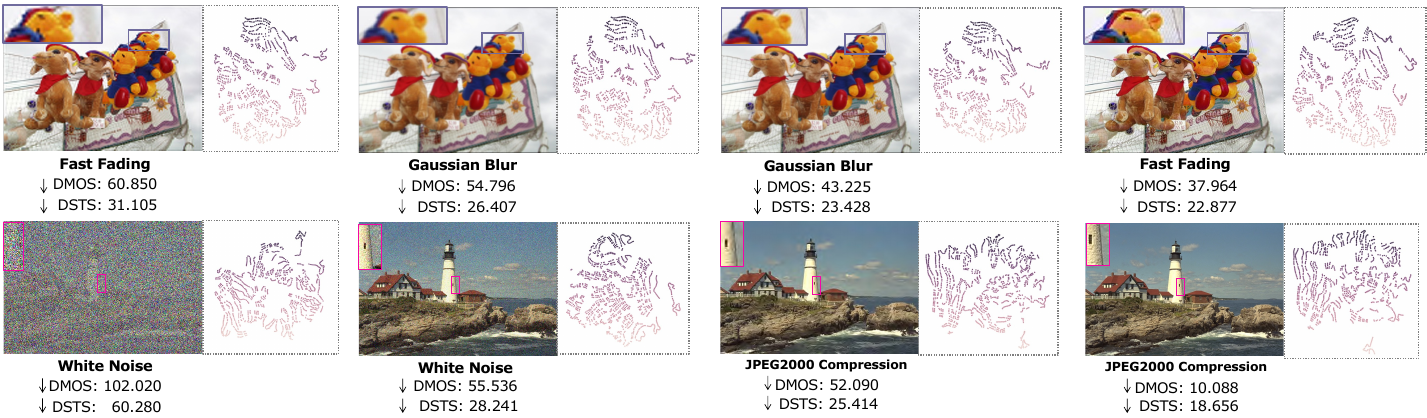}
\caption{Quality prediction results and statistical visualizations of distorted images sampled from the LIVE database. The distortion types, DMOSs, predicted DSTS values and the 2-D t-SNE visualized image inner statistical distributions are shown. }
\label{cases}
\end{figure*}

\subsection{Implementation Details}
\noindent\textbf{Backbone training}\quad To obtain the shape-texture oriented image statistics, we first train the feature extraction backbone EfficientNet-b7~\cite{tan2019efficientnet} for the image classification task on the stylized-ImageNet, which is generated following the settings in \cite{geirhos2018imagenet}. The stylized-ImageNet is ImageNet processed with style transfer algorithms, which is believed as a way of eliminating texture information in images. Additionally, we also train the shape-biased VGG16~\cite{vgg19} and ResNet50~\cite{resnet101} for comparison. We use an initial learning rate 0.01 for the EfficientNet, and 0.1 for VGG16 and ResNet50, using the exponential decay strategy with the decay parameter of 0.1 for every 10 epochs. By default, the training batch size is set to 256, the weight decay is set to 5e-6, and the momentum for stochastic gradient descent (SGD) optimization is 0.9. For the texture-biased backbones, we directly adopt the VGG16, ResNet50, and EfficientNet-b7 pretrained on ImageNet~\cite{russakovsky2015imagenet}.

\noindent\textbf{Implementation of DSTS}\quad The convolutional responses of the second to the sixth stages are employed to extract image statistics, which have 32, 48, 80, 160, and 224 channels, respectively. The image shape-texture outer statistics are determined based on a subcollection with 500 images from the DIV2K~\cite{agustsson2017ntire} database, which are fixed and subsequently applied to all the testing databases. The filtering threshold $\tau$ of the structure indicator is set to the average value of $\boldsymbol{\tilde{I}}_{\sigma}$. All models are constructed with PyTorch on a machine equipped with six NVIDIA Geforce RTX 4090 GPUs. It is worth noting that there is no content overlapping between the pristine image set and testing databases unless stated otherwise. 

\subsection{Performance Evaluation on Quality Prediction}
We compare the performance of DSTS along with its shape component DSS and texture component DTS against nine existing OU-NRIQA methods that have public-available source codes, including NIQE~\cite{niqe}, quality-aware clustering (QAC)~\cite{qac}, perception-based image quality evaluator (PIQUE)~\cite{pique}, LPSI~\cite{lpsi}, ILNIQE~\cite{ilniqe}, dipIQ~\cite{dipiq}, SNP-NIQE~\cite{snpniqe}, NPQI~\cite{npqi}, and ContentSep~\cite{babu2023no}. Besides, we also compare the generalization ability with five popular OA-BIQA models under the cross-database setting, including PaQ2PiQ~\cite{ying2020patches}, HyperIQA~\cite{su2020blindly}, MANIQA~\cite{yang2022maniqa}, VCRNet~\cite{pan2022vcrnet}, and MUSIQ~\cite{ke2021musiq}.
For performance comparisons, we follow the official codes and parameter settings described in the papers. 
\subsubsection{Performance on Each Individual Database}
\begin{table*}[!tbp]
    \centering
    \fontsize{8}{9}\selectfont
    \caption{Quality Prediction Results in Terms Of SRCC on Each Individual Distortion Type. The DSTS Performance is Highlighted in Purple, and the First and Second Best Performance are Highlighted in \textbf{Bold} and \underline{Underline} Respectively.}
    \setlength{\tabcolsep}{1.2mm}{\begin{tabular}{c|c|cccccccc|cc>{\columncolor{mypurple!20}}c}
    \toprule
    Database & Distortion Type & NIQE  & QAC   & PIQUE  & LPSI  & ILNIQE & dipIQ & SNP-NIQE & NPQI  &DSS& DTS &\textbf{DSTS} \\
    \midrule
    \multirow{24}[2]{*}{TID2013} 
    & Additive Gaussian Noise & 0.819 & 0.743 & 0.856 & 0.769 & \underline{0.877} & 0.865 & \textbf{0.886} & 0.626 &0.731& 0.850&0.852 \\
          & Additive Noise in Color Components & 0.670 & 0.718 & 0.758 & 0.496 & \textbf{0.816} & \underline{0.769} & 0.732 & 0.297 &0.679& 0.738 &0.741\\
          & Spatially Correlated Noise & 0.666 & 0.169 & 0.293 & 0.697 & \textbf{0.923} &0.809 & 0.651 & -0.233 &0.782& 0.815&\underline{0.817} \\
          & Masked Noise & 0.746 & 0.704 & 0.593 & 0.046 & 0.513 & 0.725 & 0.738 & 0.662 &0.569 &0.649&0.645 \\
          & High Frequency Noise & 0.845 & 0.863 & 0.892 & \textbf{0.925} & 0.869 & 0.864 & 0.873 & 0.821 &0.819& \underline{0.888} &0.886\\
          & Impulse Noise & 0.744 & 0.792 & \underline{0.800} & 0.432 & 0.756 & 0.788 & \textbf{0.801} & 0.575 &0.652& 0.770 &0.776\\
          & Quantization Noise & 0.850 & 0.709 & 0.751 & 0.854 & \underline{0.871} & 0.799 & 0.857 & 0.773 &0.668& 0.870&\textbf{0.872} \\
          & Gaussian Blur & 0.797 & 0.846 & 0.828 & 0.841 & 0.815 & \textbf{0.905} & 0.863 & 0.759 &\underline{0.866}& 0.861 &0.863\\
          & Image Denoising & 0.590 & 0.338 & 0.644 & -0.249 & 0.749 & 0.069 & 0.612 & 0.641 &0.851& \underline{0.875} &\textbf{0.878}\\
          & JPEG Compression & 0.843 & 0.837 & 0.793 & \textbf{0.912} & 0.834 & \underline{0.912} & 0.878 & 0.847 &0.833& 0.895 &0.895\\
          & JPEG2000 Compression & 0.889 & 0.790 & 0.854 & 0.899 & 0.858 & 0.919 & 0.881 & 0.851 &0.914& \textbf{0.933} &\underline{0.931}\\
          & JPEG Transmission Errors & 0.000 & 0.049 & 0.061 & 0.091 & 0.282 & \textbf{0.709} & 0.282 & -0.031 &\underline{0.510}& 0.424 &0.427\\
          & JPEG2000 Transmission Errors & 0.511 & 0.407 & 0.113 & \textbf{0.611} & 0.524 & 0.369 & 0.592 & -0.298 &\underline{0.606}& 0.526 &0.536\\
          & Non Eccentricity Pattern Noise & -0.069 & 0.048 & -0.010  & \underline{0.052} & -0.081 & \textbf{0.371} & 0.017 & -0.025 &-0.060& -0.016 &-0.014\\
          & Local Block-wise Distortions & -0.131 & \underline{0.247} & 0.178 & 0.137 & -0.132 & \textbf{0.291} & -0.037 & -0.072 &0.028& 0.089 &0.101\\
          & Mean Shift & -0.163 & \underline{0.306} & 0.271 & \textbf{0.341} & 0.184 & 0.084 & -0.122 & -0.090 &0.082& 0.182 &0.185\\
          & Contrast Change & -0.017 & -0.207 & -0.072 & 0.199 & 0.014 & -0.145 & 0.154 & \textbf{0.463} &\underline{0.351}& 0.284 &0.278\\
          & Change of Color Saturation & -0.246 & 0.368 & 0.268 & 0.302 & -0.165 & 0.068 & -0.107 & -0.346 &0.283& \textbf{0.511} &\underline{0.508}\\
          & Multiplicative Gaussian Noise & 0.693 & \textbf{0.790} & 0.732 & 0.696 & 0.694 & \underline{0.788} & 0.741 & 0.396 &0.583& 0.758 &0.762\\
          & Comfort Noise & 0.154 & -0.152 & -0.133 & 0.018 & \underline{0.361} & 0.359 & 0.208 & -0.345 &\textbf{0.436}& 0.299 &0.301\\
          & Lossy Compression of Noisy Images & 0.803 & 0.640 & 0.637 & 0.236 & 0.829 & \textbf{0.851} & 0.831 & 0.373 &0.800& 0.846 &\underline{0.849}\\
          & Color Quantization with Dither & 0.783 & \underline{0.873} & 0.812 & \textbf{0.900} & 0.750 & 0.756 & 0.789 & 0.756 &0.609& 0.804 &0.801\\
          & Chromatic Aberrations & 0.562 & 0.625 & 0.676 & 0.695 & 0.679 & 0.700 & 0.634 & 0.534 &\textbf{0.854}& 0.718 &\underline{0.723}\\
          & Sparse Sampling and Reconstruction & 0.834 & 0.786 & 0.823 & 0.862 & 0.864 & 0.761 & 0.828 & 0.825 &0.880& \underline{0.920} &\textbf{0.923}\\
    \midrule
    \multirow{6}[2]{*}{CSIQ} & Additive Gaussian Noise & 0.811 & 0.823 & \underline{0.901} & 0.664 & 0.851 & \textbf{0.902} & 0.876 & 0.714 &0.805& 0.831 &0.829\\
          & Gaussian Blur & 0.875 & 0.819 & 0.843 & 0.880 & 0.830 & \textbf{0.900} & \underline{0.897} & 0.864 &0.876& 0.873 &0.875\\
          & Global Contrast Decrements & 0.239 & -0.250 & 0.093 & 0.543 & 0.518 & -0.155 & 0.435 & \textbf{0.668} &\underline{0.611}& 0.496 &0.507\\
          & Additive Pink Gaussian Noise & 0.299 & -0.004 & 0.120 & 0.247 & \textbf{0.877} & -0.165 & 0.260 & 0.005 &0.729& 0.785 &\underline{0.785}\\
          & JPEG Compression & 0.862 & 0.878 & 0.824 & \textbf{0.927} & 0.876 & 0.925 & 0.906 & 0.887 &0.883& 0.912 &\underline{0.913}\\
          & JPEG2000 Compression & 0.894 & 0.865 & 0.848 & 0.900 & 0.894 & \textbf{0.936} & 0.896 & 0.877 &0.910& 0.934 &\underline{0.934}\\
    \midrule
    \multirow{5}[2]{*}{LIVE} & JPEG Compression &0.942       &0.936      &0.908       &0.968       &0.942       &\underline{0.969}       &\textbf{0.970}       &0.948       &0.940 &0.964  &0.963\\
          & JPEG2000 Compression &0.919       &0.862       &0.902       &0.930       &0.894       &0.954       &0.918       &0.888      &0.926 &\textbf{0.957}  &\underline{0.956}\\
          & White Noise &0.972       &0.951       &\textbf{0.985}       &0.956       &\underline{0.981}       &0.974       &0.978       &0.960     &0.969  &0.978  &0.978\\
          & Gaussian Blur &0.933       &0.913       &0.909       &0.916       &0.915       &0.934       &\underline{0.951}       &0.911     &0.943  &0.949  &\textbf{0.951}\\
          & Fast-fading &0.863       &0.823       &0.785       &0.781       &0.833       &0.860       &0.850       &0.837       &0.873 &\textbf{0.908}  &\underline{0.906}\\
    % \multirow{2}[2]{*}{MDLIVE} & Blur + JPEG &0.872       &0.396       &0.768       &0.839       &\textbf{0.892}       &0.698       &0.842       &0.851      & &0.787  &\\
    %       & Blur + Gaussian Noise &0.795       &0.471       &0.247       &-0.001       &\textbf{0.882}       &0.739       &0.796       &0.769       &&0.775  &\\
    \midrule
    \multirow{2}[2]{*}{MDIVL} & Blur + JPEG &0.760       &0.554       &0.658       &0.735       &0.791       &0.651       &\textbf{0.809}      &0.774       &0.778 &0.804  &\underline{0.806}\\
          & Gaussian Noise + JPEG &0.444       &0.528       &0.401       &0.469       &0.580       &0.773       &0.552       &0.537       &0.659&\textbf{0.778}  &\underline{0.776}\\      
    \bottomrule
    \end{tabular}%
    }
\label{result6}
\end{table*}
We compare the performance of DSTS with nine OU-BIQA models on IQA databases that contain artificial and authentic distortions respectively. The comparisons in terms of SRCC, PLCC and RMSE are included in Table~\ref{result1} and \ref{result2}. It can be observed from the results that DSTS obtains SOTA performance on eight out of eleven databases and achieves second-best performance on the other two databases, which indicates a competitive performance on both synthetic and authentic distortions. Besides, one can observe that DSTS shows great superiority on all the databases with authentic distortions, indicating the promising capability of capturing authentic distortions. In addition, we provide the illustration of the linear correlation between the predicted DSTS values and objective quality scores in SM, revealing DSTS possesses promising quality prediction accuracy. Furthermore, we find out that the performance of statistical comparison incorporating shape and texture bias preponderate on different databases, and a unified shape-texture statistical representation can achieve a stable and superior performance on all the databases. These experimental results demonstrate that even without the assistance of reference images and quality opinions, properly designed OU-BIQA models can still achieve a promising performance. In Fig.~\ref{cases}, we provide the differential mean opinion scores (DMOS) and predicted DSTS scores of eight distorted images corrupted from two reference images. One can observe the disturbance on image inner statistics becomes severer with the decrease of image quality, and DSTS  increases correspondingly. This reveals that DSTS is able to measure perceptual quality effectively even in a completely-blind condition.
\subsubsection{Performance on Each Individual Distortion Type}
To better investigate DSTS performance on specific distortions, we conduct  comparisons on LIVE, CSIQ, TID2013 and MDIVL. The results are listed in Table~\ref{result6}. We can observe that regarding commonly encountered distortions such as ``Gaussian Noise'', ``JPEG/JP2K Compression'', ``Gaussian Blur'', and ``Fast-fading'', DSTS delivers promising results. Especially, for the ``Image Denoising'', ``Sparse Sampling and Reconstruction'', and ``Additive Pink Gaussian Noise'' subsets, none of the compared models can achieve acceptable performance except for DSTS. This reveals the proposed DSTS is able to generalize to different distortion types because it is not dedicatedly designed for any specific distortions.

\subsubsection{Performance on Generative Distortions}
Distortions sourced from generative models possess unique characters that are challenging for current IQA models~\cite{jinjin2020pipal}. To testify the effectiveness of DSTS on generative distortions, we compare its performance with NIQE and IL-NIQE on three GAN image IQA databases. Moreover, we also focus on the different utility of texture-biased and shape-biased statistics respectively, to obtain insights of the effectiveness of biased statistics on generative distortions. The results listed in Table~\ref{gandistortion} show that the proposed DSTS achieves top-tier performance on generative distortion. Besides, we observe that shape-biased statistics contribute more to quality assessment in most cases, indicating that generative distortions disturb more shape-relevant image cues compared with texture-relevant ones.
\begin{table}[!tbp]
    \fontsize{8}{10}\selectfont
    \centering
    \caption{Quality Prediction Performance Comparisons on IQA Databases with Generative Distortions. The DSTS Performance is Highlighted in Purple. The Best Performer is highlighted in \textbf{Bold}. Cases Where Shape Outperforms Texture are Marked in {\color{myred}Red} and Cases Where Texture Outperforms Shape are Marked in {\color{mygreen}Green}.}
    \setlength{\tabcolsep}{1mm}{\begin{tabular}{cccccc}
    \toprule
    \multirow{2}{*}{OU-BIQA}  &\multicolumn{3}{c}{LGIQA} &\multirow{2}{*}{GFIQA-20k} &\multirow{2}{*}{AGIQA-3k} \\
    \cline{2-4}
    &city &cat &ffhq &&  \\
    \midrule
    NIQE &0.579&0.493&0.371&0.501&0.534\\
    IL-NIQE &0.827&0.425&\textbf{0.557}&0.714&0.594\\
    \midrule 
    DSS-VGG16 (Ours)  &\color{myred}\textbf{0.831}&0.448&\color{myred}0.482&\color{myred}0.778&\color{myred}0.688\\
    DTS-VGG16 (Ours)  &0.824&\color{mygreen}0.461&0.410&0.773&0.665\\
    \cdashline{1-6}
    DSS-ResNet50 (Ours)  &\color{myred}0.826&0.435&\color{myred}0.458&\color{myred}0.754&\color{myred}0.653\\
    DTS-ResNet50 (Ours)  &0.808&\color{mygreen}0.449&0.384&0.705&0.604\\
    \cdashline{1-6}
    DSS &\color{myred}0.825&0.483&\color{myred}0.465&\color{myred}0.865&\color{myred}\textbf{0.707}\\
    DTS &0.760&\color{mygreen}0.540&0.432&0.833&0.675\\
    \rowcolor{mypurple!20} {\textbf{DSTS}}&0.810&\textbf{0.542}&0.479&\textbf{0.843}&0.693 \\
    \bottomrule
    \end{tabular}
    }
    \label{gandistortion}
\end{table}

\subsubsection{Generalization Ability}
We compare the proposed DSTS with five OA-BIQA methods to demonstrate our superiority on generalization ability. The results are listed in Table~\ref{oabiqa}. We observe that DSTS achieves SOTA in most cases without dataset-specific adjustment, indicating it possesses a high generalization ability even compared with OA methods. Such ability is enriched by the shape-texture statistical representations, which can benefit IQA in more universal applications.    
\begin{table}[!tbp]
    \fontsize{8}{10}\selectfont
    \centering
    \caption{Quality Prediction Performance Comparisons With OA-BIQA Methods In Terms of SRCC.The DSTS Performance is Highlighted in Purple. The First and Second Best Performer is highlighted in \textbf{Bold} and \underline{Underline}.}
    \setlength{\tabcolsep}{1mm}{\begin{tabular}{cccccccccc}
    \toprule
    Method  &LIVE &CSIQ &TID2013 &KADID &MDIVL &KonIQ &CLIVE\\
    \midrule
    PaQ2PiQ &0.479&0.564&0.401&0.383&0.536&0.721&0.718\\
    HyperIQA &0.755&0.581&0.384&0.468&0.617&-&\underline{0.761}\\
    MANIQA &0.779&0.662&0.451&0.438&0.511&-&\textbf{0.840}\\
    VCRNet &-&0.681&0.512&0.444&0.475&0.606&0.557\\
    MUSIQ &0.734&0.588&0.474&0.464&0.592&-&0.722\\
    \midrule 
    DSS &0.880&0.770&0.511&0.567&0.643&0.676&0.457\\
    DTS &\textbf{0.936}&\underline{0.777}&\underline{0.536}&\underline{0.598}&\underline{0.787}&\underline{0.733}&0.482\\
    \rowcolor{mypurple!20} {\textbf{DSTS}} &\underline{0.935}&\textbf{0.798}&\textbf{0.538}&\textbf{0.622}&\textbf{0.803}&\textbf{0.753}&0.483 \\
    \bottomrule
    \end{tabular}
    }
    \label{oabiqa}
\end{table}

\section{Applications on Personalized Blind Image Quality Assessment}
\label{customized iqa}
\begin{figure}[!tbp]
    \centering
    \includegraphics[scale=0.5]{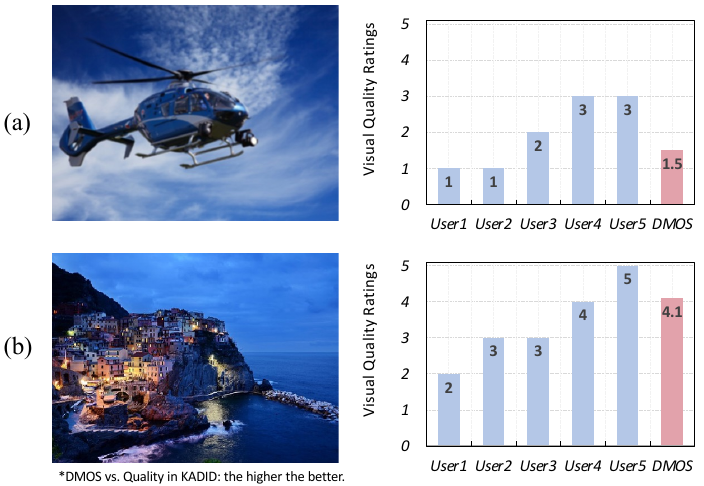}
    \caption{Exemplar distorted images and corresponding quality ratings from five individual users from the KADID database. The quality rating ranges from 1 to 5, indicating the lowest to highest perceptual quality. }
\label{personaliqa}
\end{figure}

The proposed DSTS possesses the significant capability of capturing quality characteristics of images, by unifying a shape-texture statistical description. Upon this specialty, we further extend the DSTS to the application of personalized blind image quality prediction. As shown in Fig.~\ref{personaliqa}, the uniqueness of IQA personalization is that it measures an individual's quality preference instead of the average perception characterized by MOS or DMOS. In application, for a specific user, DSTS can conduct personalized quality prediction of distorted images based on the user-tailored outer statistics. 
\begin{figure*}[!tbp]
    \centering    
    \includegraphics[scale=0.55]{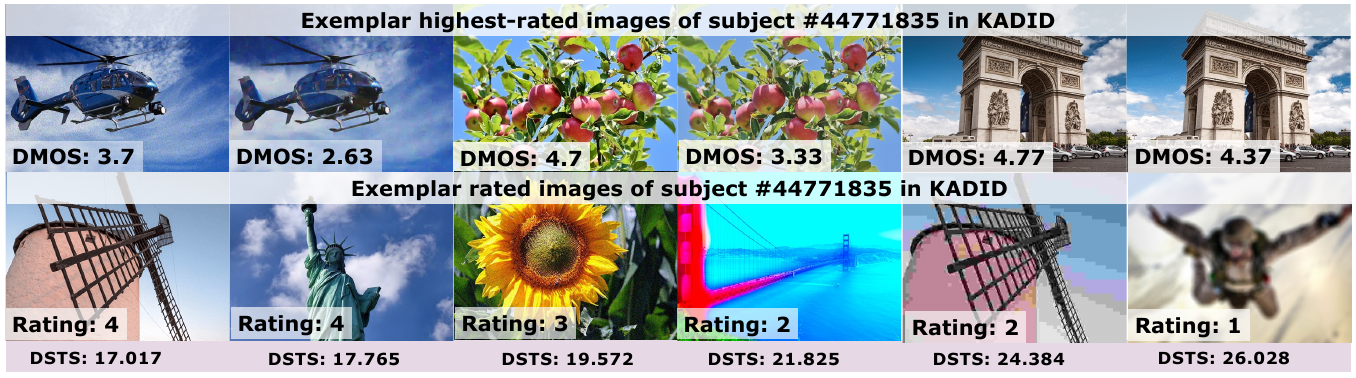}
    \caption{Exemplar images with ratings from the subject No. 44771835 of the KADID database. The first row contains the distorted images with the highest quality rating of 5, and the corresponding DMOSs are given in the bottom left corner. The second row contains another six distorted images rated by this subject from 1 to 4, ranking in the increasing order according to the predicted DSTS values. }
\label{personaliqa1}
\end{figure*}
\begin{table}[!tbp]
  \centering
  \fontsize{8}{9}\selectfont
  \caption{Comparison Results In Terms Of Average SRCC Across Images Rated By Each Subject On The Kadid Database.}
    \setlength{\tabcolsep}{3mm}{\begin{tabular}{cccc}
    \toprule
         \multicolumn{2}{c}{Method} & M>10 & M>100 \\
    \cmidrule(r){1-2} \cmidrule(r){3-4}
    \multirow{3}{*}{OA-BIQA} & BRISQUE &0.088$\pm$0.004       &0.102$\pm$0.003  \\
        &CORNIA &0.094$\pm$0.005&0.101$\pm$0.007 \\
        & CLIP-IQA &0.171$\pm$0.015       &0.181$\pm$0.020  \\
    \cline{1-4}
         \multirow{3}{*}{OU-BIQA} & NIQE  &0.108$\pm$0.002       &0.101$\pm$0.013  \\
         &IL-NIQE &0.123$\pm$0.007&0.124$\pm$0.012 \\
          & \cellcolor{mypurple!20}\textbf{DSTS}  & \cellcolor{mypurple!20}\textbf{0.352$\pm$0.003}       & \cellcolor{mypurple!20}\textbf{0.398$\pm$0.010}  \\
    \bottomrule
    \end{tabular}%
    }
  \label{personal2}%
\end{table}%
More specifically, we conduct experiments on the KADID database~\cite{kadid}, because it contains raw quality ratings from each subject of each distorted image aside from DMOSs. We consider the images labeled by the same user as a set. To achieve personalized quality assessment with DSTS, the personalized outer statistics are derived from the highest-rated images in each set, and 10 other images rated by the same subject are sampled as the test set. The performance is evaluated by the rank consistency SRCC between quality ratings and DSTS values for each subject, and the final SRCC for each IQA model is acquired by averaging the SRCCs of all subjects. Since DSTS requires a reasonable number of pristine images to describe the natural image domain statistics, we only consider the subjects with the number of highest-rated images    $M \geq 10$. Moreover, the results for the case $M>100$ are also provided for comparison. We compare DSTS with three OA-BIQA methods BRISQUE~\cite{brisque}, CORNIA~\cite{corina}, CLIP-IQA~\cite{clipiqa}, and two OU-BIQA models NIQE~\cite{niqe} and IL-NIQE~\cite{ilniqe}. The results are summarized in Table~\ref{personal2}.  
We also show distorted images rated by the same subject together with the quality ratings and predicted DSTS values in Fig.~\ref{personaliqa1}. It is apparent that DSTS successfully achieves higher personalized rank correlation with individual subject's ratings and shows superiority over other OA and OU methods, which manifests DSTS's promising capability on the application of personalized BIQA.

\section{Conclusions}
In this paper, we quest the capability of deep features produced by DNNs for quality prediction by establishing a unified shape-texture statistical representation. 
Based on this representation, we further build an OU-BIQA model DSTS by formulating the inner and outer statistics on the basis of adaptively-merged shape and texture deep features. 
By measuring the statistical distance between the inner and outer shape-texture statistics, the visual quality of distorted images can be assessed effectively without any assistant from reference images and MOSs in training. Extensive experiments provide useful evidence that DSTS's quality predictions closely align with human perception, and generalize well on both artificial and authentic distortions. It also shows superiority on personalized image quality preference prediction. The promising performance of DSTS indicates the prospective for OU-BIQA methods, and might assist to reduce the over-dependence on human-labeled data for blind image quality assessment.
\bibliographystyle{IEEEtran}
\bibliography{refs}
% \printbibliography

% \appendices

\end{document}